%% file: main.tex
\documentclass{article}
\usepackage[final, nonatbib]{neurips_2022}
\usepackage[numbers]{natbib}
\usepackage[utf8]{inputenc} 
\usepackage[T1]{fontenc}    
\usepackage{hyperref}       
\usepackage{soul}

\usepackage{booktabs}       
\usepackage{amsfonts}       
\usepackage{nicefrac}       
\usepackage{microtype}      
\usepackage{xcolor}         

\usepackage{caption}
\usepackage{subcaption}
\usepackage{graphicx}

\title{Overcoming Barriers to Skill Injection in Language Modeling: Case Study in Arithmetic}

\author{%
  Mandar Sharma\\
  Virginia Tech\\
  \texttt{mandarsharma@vt.edu}\\
  \And
  Nikhil Muralidhar\\
  Stevens Institute of Technology\\
  \texttt{nmurali1@stevens.edu}\\
  \And
  Naren Ramakrishnan\\
  Virginia Tech\\
  \texttt{naren@cs.vt.edu}\\
}

\begin{document}

\maketitle

\begin{abstract}
\input{abstract}        
\end{abstract}

\section{Introduction}
\input{introduction}

\vspace{-0.25cm}
\section{Catastrophic Forgetting in Language Models}
\input{catastrophic}

\section{Overcoming Catastrophic Forgetting}
\input{experiments}

\section{Discussion}
\input{discussion}

\input{main.bbl}
\section{Appendix}
\input{appendix}

\end{document}

%% file: abstract.tex
Through their transfer learning abilities, highly-parameterized large pre-trained language models have dominated the NLP landscape for a multitude of downstream language tasks. Though linguistically proficient, the inability of these models to incorporate the learning of non-linguistic entities (numerals and arithmetic reasoning) limits their usage for tasks that require numeric comprehension or strict mathematical reasoning. However, as we illustrate in this paper, building a general purpose language model that also happens to be proficient in mathematical reasoning is not as straight-forward as training it on a numeric dataset. In this work, we develop a novel framework that enables language models to be mathematically proficient while retaining their linguistic prowess. Specifically, we offer information-theoretic interventions to overcome the catastrophic forgetting of linguistic skills that occurs while injecting non-linguistic skills into language models. 

%% file: introduction.tex
\vspace{-0.25cm}
Numerals grant objectivity to language \cite{porter}, thus, their incorporation to language models, among other abilities, allows for better information extraction and language inference \cite{madaan, naik, datatotext}. With the rise of the Math-NLP niche, several notable publications have investigated the deficiency of inherent numerical skills induced in large language models through unsupervised training \cite{wallace,johnson,zhang}. Several more offer architectures and training schemes to induce this skill through supervision \cite{spith, jiang, geva, sundar}. However, the overarching goal should remain to build numerically-capable language models that still do what they were intended to do - \textit{model language}. 

Injecting non-linguistic skills often comes at the cost of losing linguistic skills. Akin to all neural models, language models are susceptible to catastrophic forgetting \cite{kirkpatrick} when trained for multi-task learning, especially when the two tasks are substantially different - linguistic vs non-linguistic. Here, we first investigate the nature of catastrophic forgetting in the context of language models through an information-theoretic lens and subsequently establish interventions that prevent this phenomenon. Our models perform substantially better arithmetic while retaining their linguistic prowess.

%% file: catastrophic.tex
\vspace{-0.25cm}
\begin{figure*}[t]
    \centering
    \begin{subfigure}[b]{0.475\textwidth}   
        \centering 
        \includegraphics[scale=0.405]{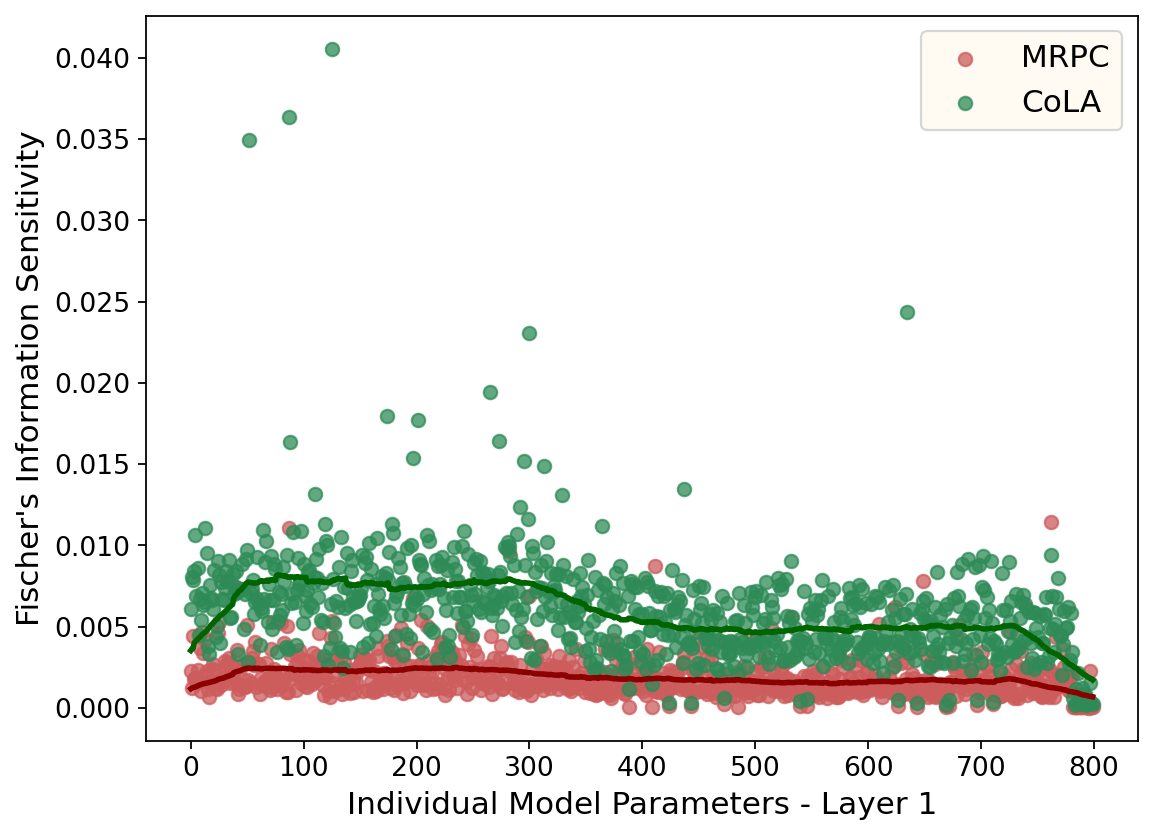}
        \caption{\footnotesize Information Sensitivity \\ Self-attention Layer-1}    
        \label{fig:fs_layer1}
    \end{subfigure}
    \hfill
    \begin{subfigure}[b]{0.475\textwidth}   
        \centering 
        \includegraphics[scale=0.405]{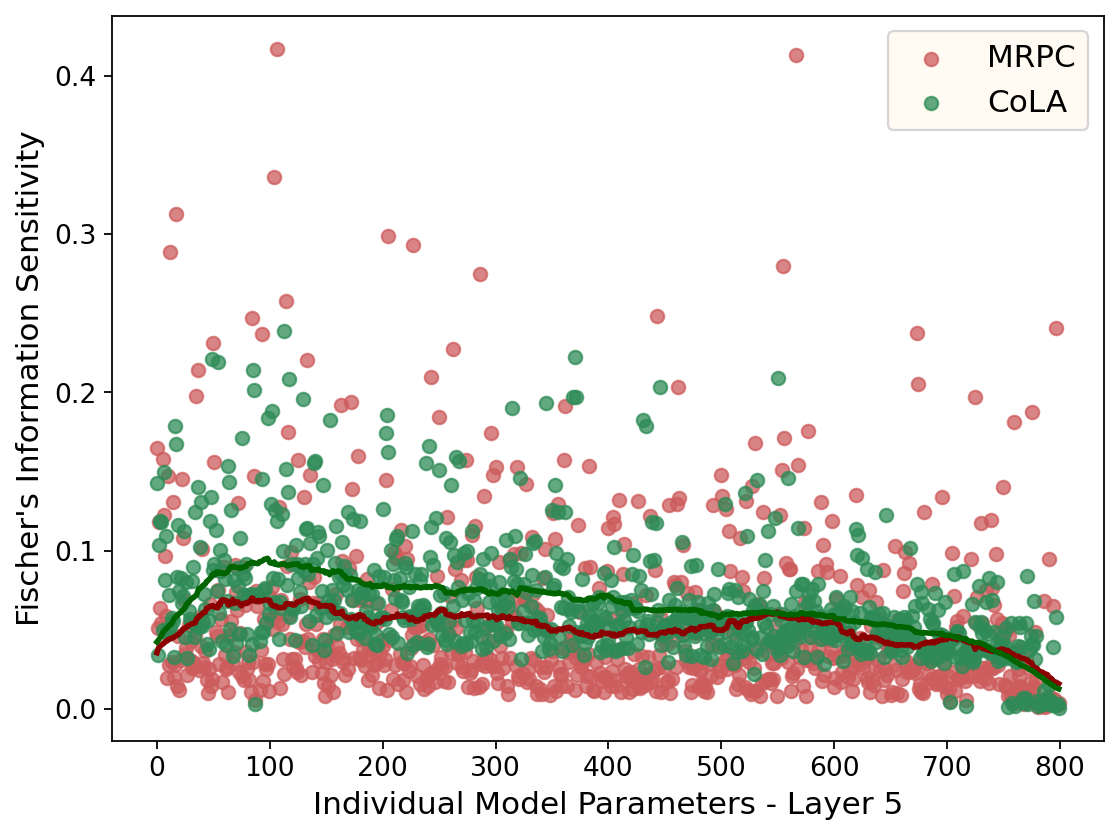}
        \caption{\footnotesize Information Sensitivity \\ Self-attention Layer-5}    
        \label{fig:fs_layer5}
    \end{subfigure}
    \vskip\baselineskip
    \begin{subfigure}[b]{0.475\textwidth}
        \centering
        \includegraphics[scale=0.55]{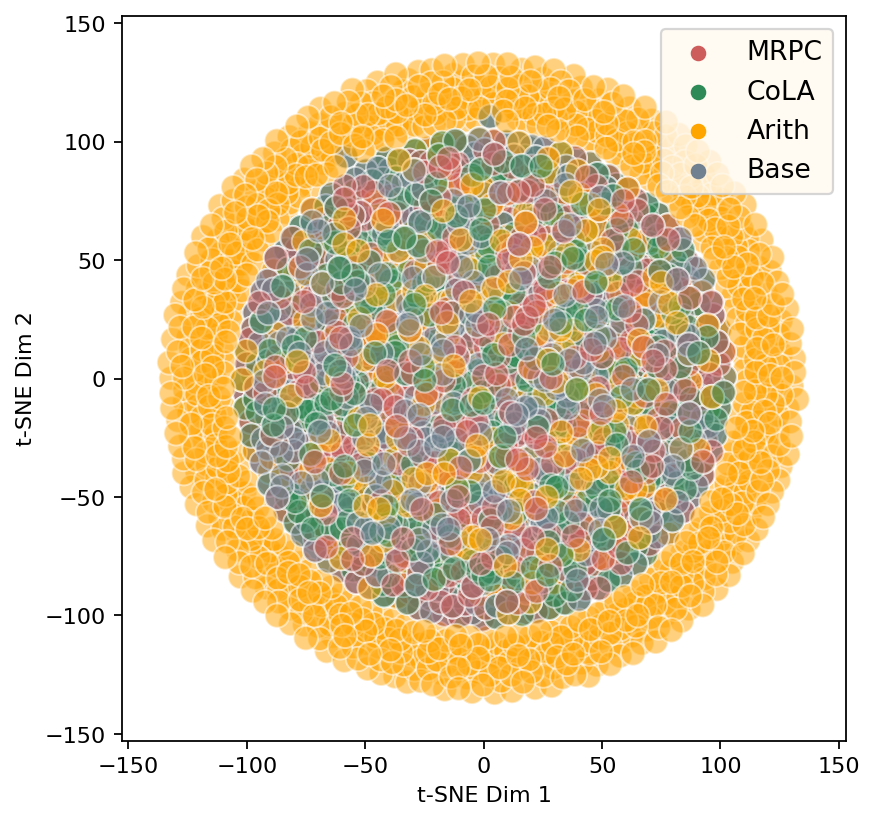}
        \caption{\footnotesize Encoder Parameter Space \\ Self-attention Layer-1}  
        \label{fig:nn_layer1}
    \end{subfigure}
    \hfill
    \begin{subfigure}[b]{0.475\textwidth}  
        \centering 
        \includegraphics[scale=0.55]{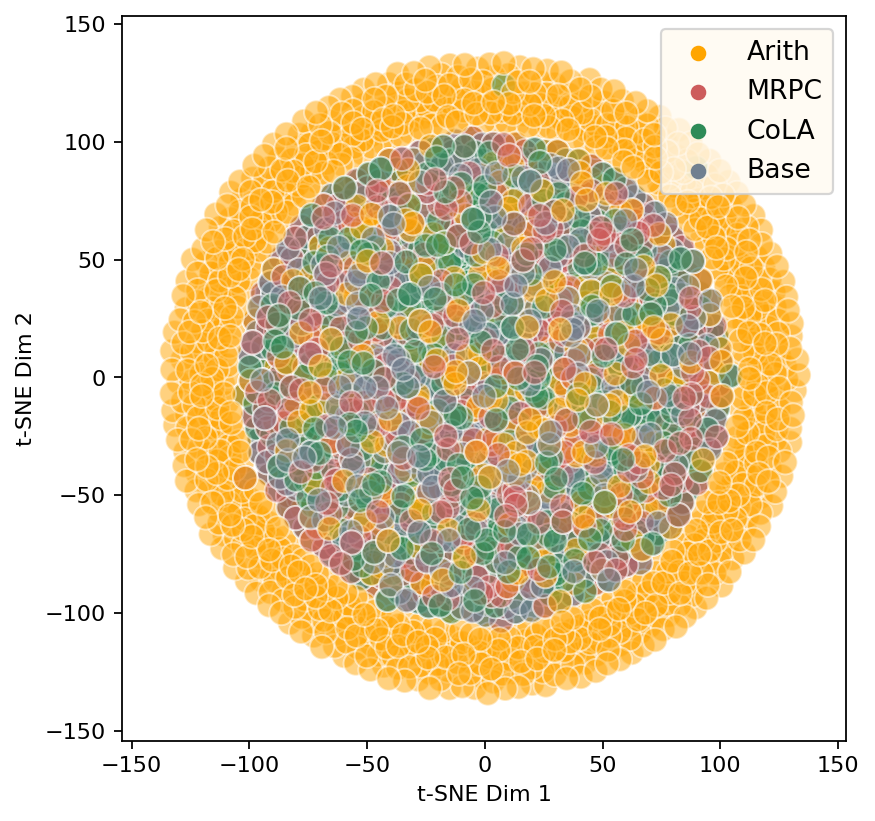}
        \caption{\footnotesize Encoder Parameter Space \\ Self-attention Layer-5}    
        \label{fig:nn_layer5}
    \end{subfigure}
    \caption{Parameter sensitivity (a \& b) for the $n=800$ most vital parameters of the encoder layers 1 \& 5 for CoLA and MRPC. The higher sensitivity of these parameters to CoLA vs MRPC lends an information-theoretic explanation of task-specific performance degradation. 2D t-SNE visualizations of the parameterization space (c \& d) for the same models \& layers - the distinction between the parameterization space of models trained for Arithmetic task vs Linguistic tasks offers another perspective into degradation of lingusitic performance when trained on non-linguistic task.}
    \label{fig:main}
    \vspace{-0.7cm}
\end{figure*}

In multi-task learning, when a model that is pre-trained on task $A$ is subsequently trained on task $B$, the weights in the model that are vital for task $A$ adapt their values to meet the objectives of task $B$. It has been found that this frequently leads to performance degradation on task $A$. This phenomenon is referred to as catastrophic forgetting \cite{kirkpatrick}. While it has been found that off-the-self pre-trained language models do establish a notion of numeric scale from their unsupervised pre-training, these are not quite enough to perform commonsense reasoning \cite{zhang}. Furthermore, without additional interventions, these notions of numeric scales fail to extrapolate for numerals outside of the training set \cite{wallace}. To meet the numerical comprehension standards required for commonsense reasoning, these model have to be trained further on numeric datasets, often resulting in the loss of their original linguistic prowess through catastrophic forgetting.

Interestingly, for langauge models, the catastrophic forgetting tendencies are not evenly spread across standardized GLUE \cite{glue} tasks - for instance, a base DistilBERT model \cite{distilbert} trained for arithmetic computation faces significant degradation in its grammatical prowess (CoLA \cite{cola}) while still retaining much of its ability for semantic equivalence (MRPC \cite{mrpc}), as seen in the second row of table \ref{table:main}. 

In an effort to better understand this phenomenon, we adopt a two-pronged approach. The analysis of the weights and biases of a converged model across multiple datasets provides a parameterization-standpoint perspective towards understanding their performance across shared tasks \cite{han,unter}. Thus, we visualize the parameterization space for models trained independently on each of the aforementioned three tasks through figures \ref{fig:nn_layer1} and \ref{fig:nn_layer5}. Through the low-dimensional t-SNE \cite{tsne} projections, we observe that the model parameterizations live in different spaces when the tasks are linguistic (CoLA and MRPC) vs non-linguistic (Arithmetic). This ties back to the premise of catastrophic forgetting - where the parameters of the DistilBERT model trained for arithmetic computation have re-adjusted their weights to a space that does not comply well with linguistic tasks. These new parameter distributions for arithmetic computation (orange) are consistent across the transformer encoder self-attention layers - as seen for layer 1 in figure \ref{fig:nn_layer1} and layer 5 in figure \ref{fig:nn_layer5}. These two figures offer the first perspective on linguistic vs non-linguistic performance degradation from a model parameterization stand-point. 

For a complementary perspective, we zoom into the contributions of the individual model parameters for our set of shared tasks through an information-theoretic lens. For a single sample $Y$ drawn from a distribution with probability desnity $f(y;\theta)$, the Fisher information index $I(\theta)$ (\ref{fisher}) quantifies the sensitivity of the parameter $\theta$ to the data instance $Y$. After selecting the top $n=800$ most vital parameters for each encoder layer for the Arithmetic task based on their Fisher information scores, we observe that the same model parameters are more sensitive to the CoLA task than to the MRPC task, as shown in figures \ref{fig:fs_layer1} and \ref{fig:fs_layer5}, again, based on their respective task-specific sensitivity scores. Thus, offering a complementary information-theoretic perspective on task-specific performance degradation.

\begin{equation}
\label{fisher}
    I(\theta) = E(\frac{d \log_{} f(Y;\theta)}{d\theta})^{2} = -E(\frac{d^{2} \log_{} f(Y;\theta)}{d\theta^{2}}) 
\end{equation}

%% file: experiments.tex
\subsection{Elastic Weight Consolidation for Language Modeling}
System-level consolidation often consists of stitching-together an amalgamated dataset that consistutes of multiple-shared tasks \cite{kumaran}. However, for general-purpose language models, the range of possible downstream tasks are so diverse that the paradigm has remained to build large models that hold linguistic prowess and are intended to be fine-tuned on a single downstream task \cite{distilbert, bert, gpt2}. Thus, being more suited to a continual learning paradigm. As these models, by their stature, are highly parameterized - it can be ascertained that there is a solution for task $B$, $\theta_{B}$, that is proximal to the solution space for task $A$, $\theta_{A}$. For such continual learning, Elastic Weight Consolidation (EWC) \cite{kirkpatrick} regularizes learning on specific network weights based on their importance to previously seen tasks, ensuring $\theta_{B}$ remains proximal to $\theta_{A}$. 

\begin{figure}[h]
    \centering
    \includegraphics[scale=0.3]{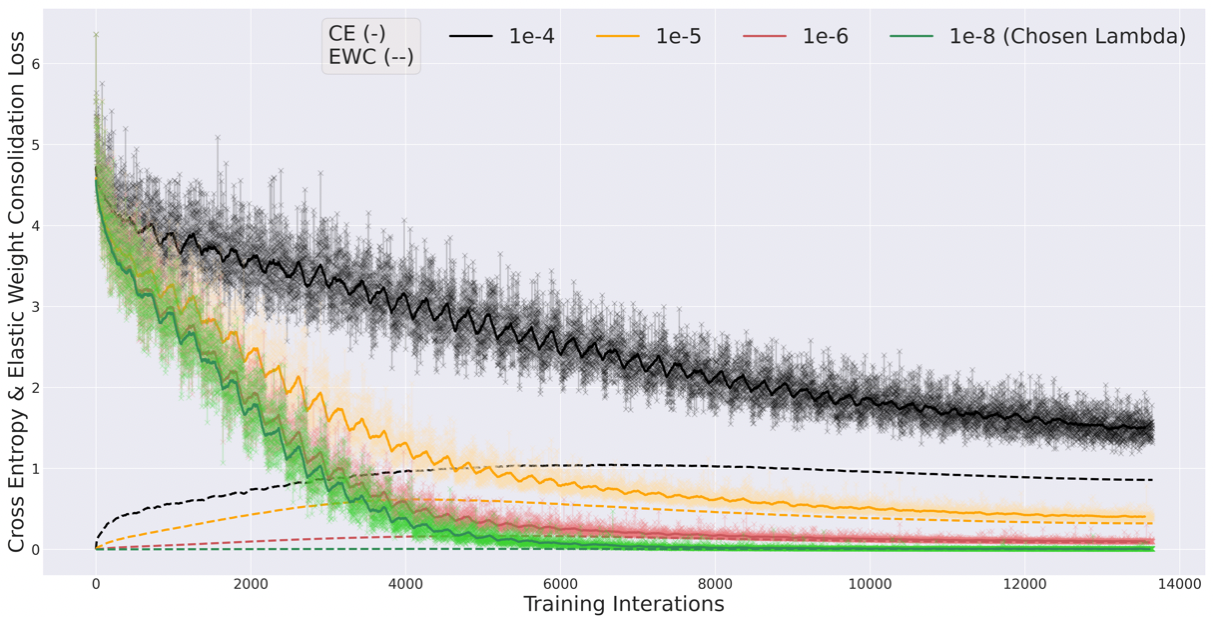}
    \caption{\footnotesize Interplay between the CE \& the EWC loss (color-matched) as a function of training iterations. The first sign of model convergence is observed at $\lambda$=1e-8, and although the convergence improves with decreasing values of $\lambda$, we choose $\lambda$ to 1e-8 to promote balanced learning of $\theta_{arith}$ with proper retention of $\theta_{gen}$.}%
    \label{fig:lambda}
\end{figure}

Thus, with the posterior distribution of the DistilBERT model approximated through its Fischer information matrix $F$ and Gaussian distribution mean from parameters $\theta^{*}_{gen}$, we train the model to predict the correct tokens representing the results of arithmetic operations through masked-modeling loss $\mathcal{L}_{arith}$ such that changes to any model parameter $i$ crucial to then general functionality of the model $\theta_{gen}$ is penalized through a quadratic penalty scaled by $\lambda$ (\ref{loss}). The selection of the hyperparameter $\lambda$ is crucial as it dictates both model convergence and balances the learning of $\theta_{arith}$ with $\theta_{gen}$. Thus, we gauge the sensitivity of model convergence with respect to $\lambda$ with a hyperparameter sweep raning from $\lambda$=1e-4 to 5e-11. Figure \ref{fig:lambda} shows the interplay between the weight consolidation loss $EWC$ and the Cross-Entropy loss for learning arithmetic $CE$ (color-matched) for different values of $\lambda$. We observe the first sign of model convergence at $\lambda$=1e-8, and although the model convergence improves with decreasing values of $\lambda$, we set $\lambda$ to 1e-8 for our experiments to promote balanced learning of $\theta_{arith}$ with proper retention of $\theta_{gen}$.

\begin{equation}
\label{loss}
    \mathcal{L}(\theta) = \mathcal{L}_{arith}(\theta) + \sum_{i} \frac{\lambda}{2} \ F_{i}(\theta_{i} - \theta_{gen,i}^{*})^{2}
\end{equation}

\subsection{Experimental Setup and Results}

Our in-house arithmetic dataset comprises of numerals modeled after the numeral distributions in real-world datasets DROP \cite{drop} and EQAUTE \cite{equate} (\textsection Appendix 5.2) with 21,838 instances of arithmetic computations. All models are trained for 50 epochs where the datasets are processed through the standard sub-word tokenization scheme of BERT \cite{bert}. These models are then fine-tuned and evaluated on the range of GLUE tasks \cite{glue}: 
\begin{itemize}
    \item Single Sentence Tasks: Corpus of Linguistic Acceptability (CoLA) \cite{cola} for grammatical fidelity with Matthews correlation coefficient as the metric and the Stanford Sentiment Treebank (SST-2) \cite{sst2} for sentiment analysis with accuracy as the metric.
    \item Similarity and Paraphrasing Tasks: The Microsoft Research Paraphrase Corpus (MRPC) \cite{mrpc} for semantic equivalence with averaged accuracy and f1 scores as the metric and the Semantic Textual Similarity Benchmark (STS-B) \cite{stsb} for sentence similarity with the Spearman rank correlation coefficient as the metric.
    \item Inference Task: The Multi-genre Natural Language Inference Corpus (MNLI) \cite{mnli} for textual entailment between a given premise and hypothesis with accuracy as the metric.
\end{itemize}

The results presented in table \ref{table:main} represent task-metrics for the models as $\mu_{\sigma}$ where $\mu$ represents the mean value and $\sigma$ represents the standard deviation across two runs with different seed values for random initialization of the model weights. Please note that the base DistilBERT model has been used off-the-shelf and thus has no standard deviation across its runs. 

\begin{table}[!h]
\centering
\footnotesize
\caption{\footnotesize{Comparative analysis between the base model, the base model trained on arithmetic, and our model. The results of the arithmetic computations are measured based on the \textit{ln} RMSE of the model output and the ground-truth numeral. The results of the GLUE tasks follow the metrics as described in \textsection 3.1.}}
\label{table:main}
\begin{tabular}{l|l|lllll}
\hline
                   & \textit{ln} RMSE          & CoLA                       & MNLI                                & MRPC                        & SST-2                      & STS-B                      \\ \hline
Base               & $3.5367_{0.000}$            & \multicolumn{1}{c}{$\textbf{0.4827}_{0.000}$} & \multicolumn{1}{c}{$\textbf{0.8074}_{0.000}$} & \multicolumn{1}{c}{$\textbf{0.8797}_{0.000}$} & \multicolumn{1}{c}{$\textbf{0.8967}_{0.000}$} & \multicolumn{1}{c}{$\textbf{0.8740}_{0.000}$} \\
Base + Arithmetics & $0.4443_{0.011}$ & $0.0000_{0.000}$                        & $0.3553_{0.001}$                   & $0.7524_{0.000}$           & $0.8761_{0.003}$          & $0.3998_{0.079}$          \\
Ours               & $\textbf{0.4360}_{0.016}$ & $0.4193_{0.000}$          & $0.7951_{0.004}$                   & $0.8570_{0.000}$              & $0.8962_{0.008}$          & $0.8626_{0.005}$          \\ \hline
\end{tabular}
\end{table}

%% file: discussion.tex
From the results in table \ref{table:main}, we infer that our model, while performing the best in arithmetic computation - orders of magnitude better than the base model, closely competes with the linguistic prowess of the base model. Besides preventing performance degradation on the MNLI, STS-B, and the MNLI tasks, for the CoLA task that suffered significant degradation with its Matthews correlation coefficient dropping to 0 after being trained on the arithmetic dataset, our model was able to revitalize it to a value competitive to the base model.

Returning to our original premise, this paper serves as a proof-of-concept that the inherent barries to skill injection caused by the catastrophic forgetting tendencies of large networks can be overcome with weight consolidation schemes specifically tailored for language modeling. Please see Appendix \textsection 5.1 for a disucssion on the limitations of our efforts. Our datasets as well as the codebase is hosted at \url{https://github.com/Mandar-Sharma/OvercomingBarriers}.

%% file: appendix.tex
\subsection{Limitations}

We believe this work to be a proof-of-concept for addressing the important problem of non-linguistic skill injection in language modeling. Besides mathematical reasoning, the set of non-linguistic skills can expand to encompass logical inference, and dataset comprehension. Additionally, our in-house arithmetic dataset comprises of addition and subtraction - we hope to extend our highly generic proposed framework to incorporate other mathematical operations like multiplication, division, and exponentiation, in addition to incorporating decimal-point numerals in the dataset.

\pagebreak
\subsection{Numeral Distribution of the Arithmetic Dataset}

\begin{figure*}[!h]
    \centering
    \begin{subfigure}[b]{0.475\textwidth}
        \centering
        \includegraphics[scale=0.15]{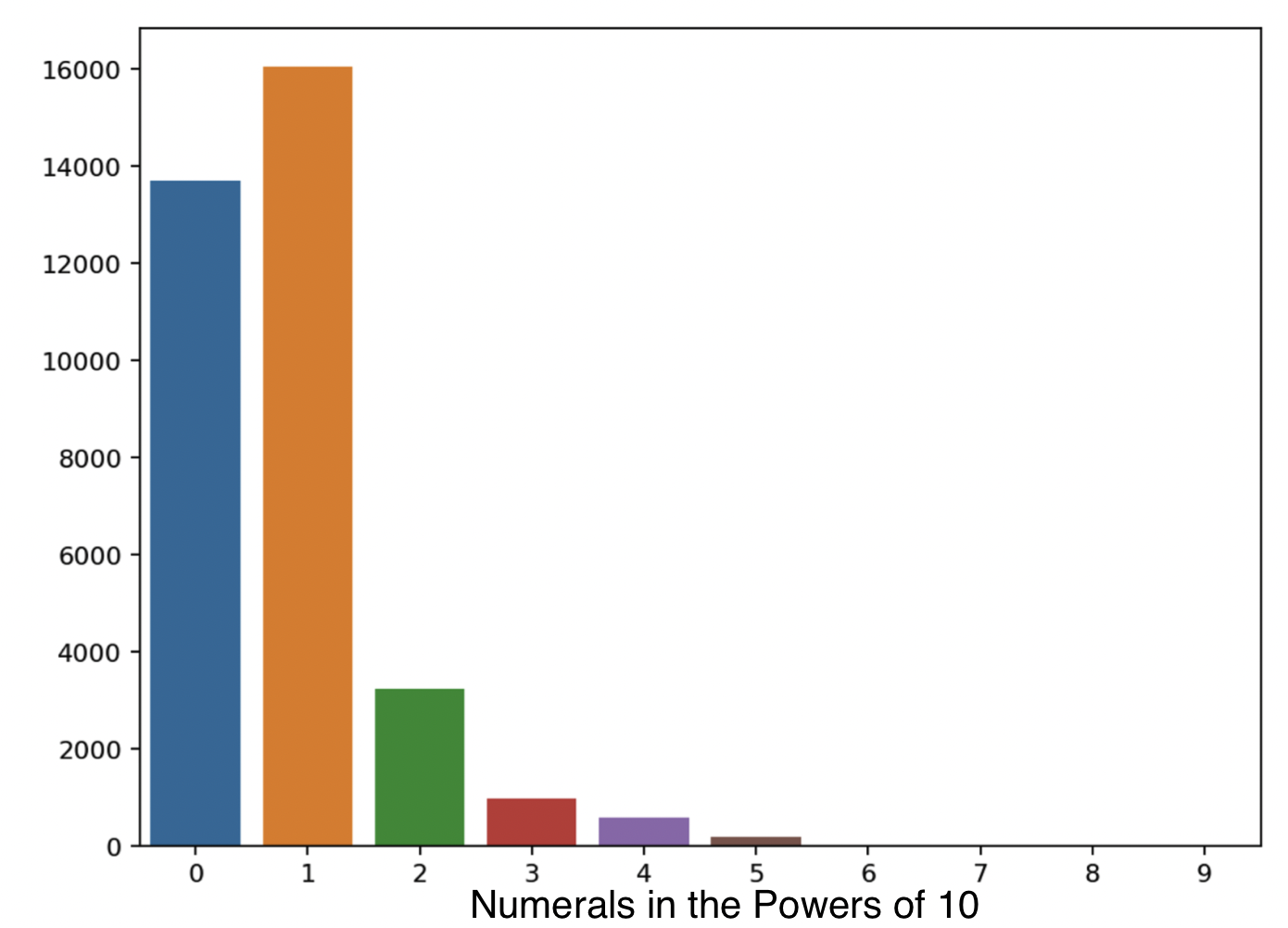}
        \caption{\footnotesize EQUATE}  
        \label{fig:equate}
    \end{subfigure}
    \hfill
    \begin{subfigure}[b]{0.475\textwidth}  
        \centering 
        \includegraphics[scale=0.285]{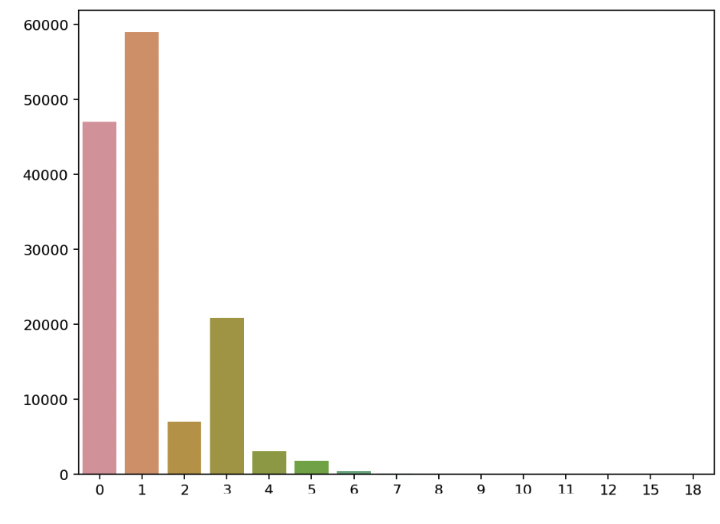}
        \caption{\footnotesize DROP}    
        \label{fig:drop}
    \end{subfigure}
    \caption{Numeral distribution histograms of the EQUATE \& DROP datasets based on powers of 10.} 
    \label{fig:dist}
\end{figure*}

%% file: main.bbl
\begin{thebibliography}{10}

\bibitem{gpt2}
Tom Brown, Benjamin Mann, Nick Ryder, Melanie Subbiah, Jared~D Kaplan, Prafulla
  Dhariwal, Arvind Neelakantan, Pranav Shyam, Girish Sastry, Amanda Askell,
  et~al.
\newblock Language models are few-shot learners.
\newblock {\em Advances in Neural Information Processing Systems},
  33:1877--1901, 2020.

\bibitem{stsb}
Daniel Cera, Mona Diabb, Eneko Agirrec, Inigo Lopez-Gazpioc, Lucia Speciad, and
  Basque~Country Donostia.
\newblock Semeval-2017 task 1: Semantic textual similarity multilingual and
  cross-lingual focused evaluation.

\bibitem{mrpc}
Bill Dolan and Chris Brockett.
\newblock Automatically constructing a corpus of sentential paraphrases.
\newblock In {\em Third International Workshop on Paraphrasing (IWP2005)},
  2005.

\bibitem{drop}
Dheeru Dua, Yizhong Wang, Pradeep Dasigi, Gabriel Stanovsky, Sameer Singh, and
  Matt Gardner.
\newblock Drop: A reading comprehension benchmark requiring discrete reasoning
  over paragraphs.
\newblock In {\em Proceedings of the 2019 Conference of the North American
  Chapter of the Association for Computational Linguistics: Human Language
  Technologies, Volume 1 (Long and Short Papers)}, pages 2368--2378, 2019.

\bibitem{geva}
Mor Geva, Ankit Gupta, and Jonathan Berant.
\newblock Injecting numerical reasoning skills into language models.
\newblock In {\em Proceedings of the 58th Annual Meeting of the Association for
  Computational Linguistics}, pages 946--958, 2020.

\bibitem{han}
Song Han, Jeff Pool, John Tran, and William Dally.
\newblock Learning both weights and connections for efficient neural network.
\newblock {\em Advances in Neural Information Processing Systems}, 28, 2015.

\bibitem{jiang}
Chengyue Jiang, Zhonglin Nian, Kaihao Guo, Shanbo Chu, Yinggong Zhao, Libin
  Shen, and Kewei Tu.
\newblock Learning numeral embedding.
\newblock In {\em Findings of the Association for Computational Linguistics:
  EMNLP 2020}, pages 2586--2599, 2020.

\bibitem{johnson}
Devin Johnson, Denise Mak, Andrew Barker, and Lexi Loessberg-Zahl.
\newblock Probing for multilingual numerical understanding in transformer-based
  language models.
\newblock In {\em Proceedings of the Third BlackboxNLP Workshop on Analyzing
  and Interpreting Neural Networks for NLP}, pages 184--192, 2020.

\bibitem{bert}
Jacob Devlin Ming-Wei~Chang Kenton and Lee~Kristina Toutanova.
\newblock Bert: Pre-training of deep bidirectional transformers for language
  understanding.
\newblock In {\em Proceedings of NAACL-HLT}, pages 4171--4186, 2019.

\bibitem{kirkpatrick}
James Kirkpatrick, Razvan Pascanu, Neil Rabinowitz, Joel Veness, Guillaume
  Desjardins, Andrei~A Rusu, Kieran Milan, John Quan, Tiago Ramalho, Agnieszka
  Grabska-Barwinska, et~al.
\newblock Overcoming catastrophic forgetting in neural networks.
\newblock {\em Proceedings of the National Academy of Sciences},
  114(13):3521--3526, 2017.

\bibitem{kumaran}
Dharshan Kumaran, Demis Hassabis, and James~L McClelland.
\newblock What learning systems do intelligent agents need? complementary
  learning systems theory updated.
\newblock {\em Trends in Cognitive Sciences}, 20(7):512--534, 2016.

\bibitem{madaan}
Aman Madaan, Ashish Mittal, Ganesh Ramakrishnan, Sunita Sarawagi, et~al.
\newblock Numerical relation extraction with minimal supervision.
\newblock In {\em Proceedings of the AAAI Conference on Artificial
  Intelligence}, volume~30, 2016.

\bibitem{sst2}
Bryan McCann, James Bradbury, Caiming Xiong, and Richard Socher.
\newblock Learned in translation: Contextualized word vectors.
\newblock {\em Advances in neural information processing systems}, 30, 2017.

\bibitem{naik}
Aakanksha Naik, Abhilasha Ravichander, Norman Sadeh, Carolyn Rose, and Graham
  Neubig.
\newblock Stress test evaluation for natural language inference.
\newblock In {\em Proceedings of the 27th International Conference on
  Computational Linguistics}, pages 2340--2353, 2018.

\bibitem{porter}
Theodore~M Porter.
\newblock Trust in numbers.
\newblock In {\em Trust in Numbers}. Princeton University Press, 1996.

\bibitem{equate}
Abhilasha Ravichander, Aakanksha Naik, Carolyn Rose, and Eduard Hovy.
\newblock Equate: A benchmark evaluation framework for quantitative reasoning
  in natural language inference.
\newblock In {\em Proceedings of the 23rd Conference on Computational Natural
  Language Learning (CoNLL)}, pages 349--361, 2019.

\bibitem{distilbert}
Victor Sanh, Lysandre Debut, Julien Chaumond, and Thomas Wolf.
\newblock Distilbert, a distilled bersion of bert: Smaller, faster, cheaper and
  lighter.
\newblock 2019.

\bibitem{datatotext}
Mandar Sharma, Ajay Gogineni, and Naren Ramakrishnan.
\newblock Innovations in neural data-to-text generation.
\newblock {\em arXiv preprint arXiv:2207.12571}, 2022.

\bibitem{spith}
Georgios Spithourakis and Sebastian Riedel.
\newblock Numeracy for language models: Evaluating and improving their ability
  to predict numbers.
\newblock In {\em Proceedings of the 56th Annual Meeting of the Association for
  Computational Linguistics (Volume 1: Long Papers)}, pages 2104--2115, 2018.

\bibitem{sundar}
Dhanasekar Sundararaman, Shijing Si, Vivek Subramanian, Guoyin Wang, Devamanyu
  Hazarika, and Lawrence Carin.
\newblock Methods for numeracy-preserving word embeddings.
\newblock In {\em Proceedings of the 2020 Conference on Empirical Methods in
  Natural Language Processing (EMNLP)}, pages 4742--4753, 2020.

\bibitem{unter}
Thomas Unterthiner, Daniel Keysers, Sylvain Gelly, Olivier Bousquet, and Ilya
  Tolstikhin.
\newblock Predicting neural network accuracy from weights.
\newblock 2020.

\bibitem{tsne}
Laurens Van~der Maaten and Geoffrey Hinton.
\newblock Visualizing data using t-sne.
\newblock {\em Journal of Machine Learning Research}, 9(11), 2008.

\bibitem{wallace}
Eric Wallace, Yizhong Wang, Sujian Li, Sameer Singh, and Matt Gardner.
\newblock Do nlp models know numbers? probing numeracy in embeddings.
\newblock In {\em Proceedings of the 2019 Conference on Empirical Methods in
  Natural Language Processing and the 9th International Joint Conference on
  Natural Language Processing (EMNLP-IJCNLP)}, pages 5307--5315, 2019.

\bibitem{glue}
Alex Wang, Amanpreet Singh, Julian Michael, Felix Hill, Omer Levy, and Samuel
  Bowman.
\newblock Glue: A multi-task benchmark and analysis platform for natural
  language understanding.
\newblock In {\em Proceedings of the 2018 EMNLP Workshop BlackboxNLP: Analyzing
  and Interpreting Neural Networks for NLP}, pages 353--355, 2018.

\bibitem{cola}
Alex Warstadt, Amanpreet Singh, and Samuel~R Bowman.
\newblock Cola: The corpus of linguistic acceptability.
\newblock 2019.

\bibitem{mnli}
Adina Williams, Nikita Nangia, and Samuel Bowman.
\newblock A broad-coverage challenge corpus for sentence understanding through
  inference.
\newblock In {\em Proceedings of the 2018 Conference of the North American
  Chapter of the Association for Computational Linguistics: Human Language
  Technologies, Volume 1 (Long Papers)}, pages 1112--1122, 2018.

\bibitem{zhang}
Xikun Zhang, Deepak Ramachandran, Ian Tenney, Yanai Elazar, and Dan Roth.
\newblock Do language embeddings capture scales?
\newblock In {\em Proceedings of the Third BlackboxNLP Workshop on Analyzing
  and Interpreting Neural Networks for NLP}, pages 292--299, 2020.

\end{thebibliography}
